\documentclass[10pt,twocolumn,letterpaper]{article}

\usepackage{cvpr}              
\definecolor{cvprblue}{rgb}{0.21,0.49,0.74}
\usepackage[pagebackref,breaklinks,colorlinks,allcolors=cvprblue]{hyperref}
\usepackage{graphicx}
\usepackage{amsmath}
\usepackage{xcolor}
\usepackage{pifont}
\usepackage{booktabs}
\usepackage{multirow}

\def\paperID{5093} 
\def\confName{CVPR}
\def\confYear{2026}

\title{YOLOA: Real-Time Affordance Detection via LLM Adapter}

\author{
	Yuqi Ji$^{1}$\footnotemark[1], Junjie Ke$^{2}$\footnotemark[1], Lihuo He$^{1}$\footnotemark[2], Jun Liu$^{1}$, Kaifan Zhang$^{1}$,Yu-Kun Lai$^{3}$, Guiguang Ding$^{2}$, Xinbo Gao$^{1}$\\[2pt]
	$^1$\textit{School of Electronic Engineering, Xidian University}\quad
	$^2$\textit{School of Software, Tsinghua University} \\
	$^3$\textit{School of Computer Science and Informatics, Cardiff University} \\
	\small
	\texttt{\{jyq,lljun,kaifanzhang\}@stu.xidian.edu.cn,}\quad
	\texttt{\{lhhe\footnotemark[2],xbgao\}@mail.xidian.edu.cn,}\\
	\small
	\texttt{jjke@mail.tsinghua.edu.cn,}\quad
	\texttt{dinggg@tsinghua.edu.cn,}\quad
	\texttt{Yukun.Lai@cs.cardiff.ac.uk}
}

\begin{document}
\maketitle
\renewcommand{\thefootnote}{\fnsymbol{footnote}}
\footnotetext[1]{Equal contribution.} 
\footnotetext[2]{Corresponding author.}
\begin{abstract}
Affordance detection aims to jointly address the fundamental  ``what–where–how''  challenge in embodied AI by understanding ``what'' an object is, ``where" the object is located, and ``how'' it can be used. However, most affordance learning methods focus solely on ``how''   objects can be used while neglecting the ``what'' and ``where'' aspects. Other affordance detection methods treat object detection and affordance learning as two independent tasks, lacking effective interaction and real-time capability. To overcome these limitations, we introduce YOLO Affordance (YOLOA), a real-time affordance detection model that jointly handles these two tasks via a large language model (LLM) adapter. Specifically, YOLOA employs a lightweight detector consisting of object detection and affordance  learning branches refined through the LLM Adapter. During training, the LLM Adapter interacts with object and affordance preliminary predictions to refine both branches by generating more accurate class priors, box offsets, and affordance gates.  Experiments on our relabeled ADG-Det and IIT-Heat benchmarks demonstrate that YOLOA achieves state-of-the-art accuracy (52.8 / 73.1 mAP on ADG-Det / IIT-Heat) while maintaining real-time performance (up to 89.77 FPS, and up to 846.24 FPS for the lightweight variant). This indicates that YOLOA achieves an excellent trade-off between accuracy and efficiency.
\end{abstract}    
\section{Introduction}

\begin{figure}[!t]
	\centering
	\includegraphics[width=\columnwidth]{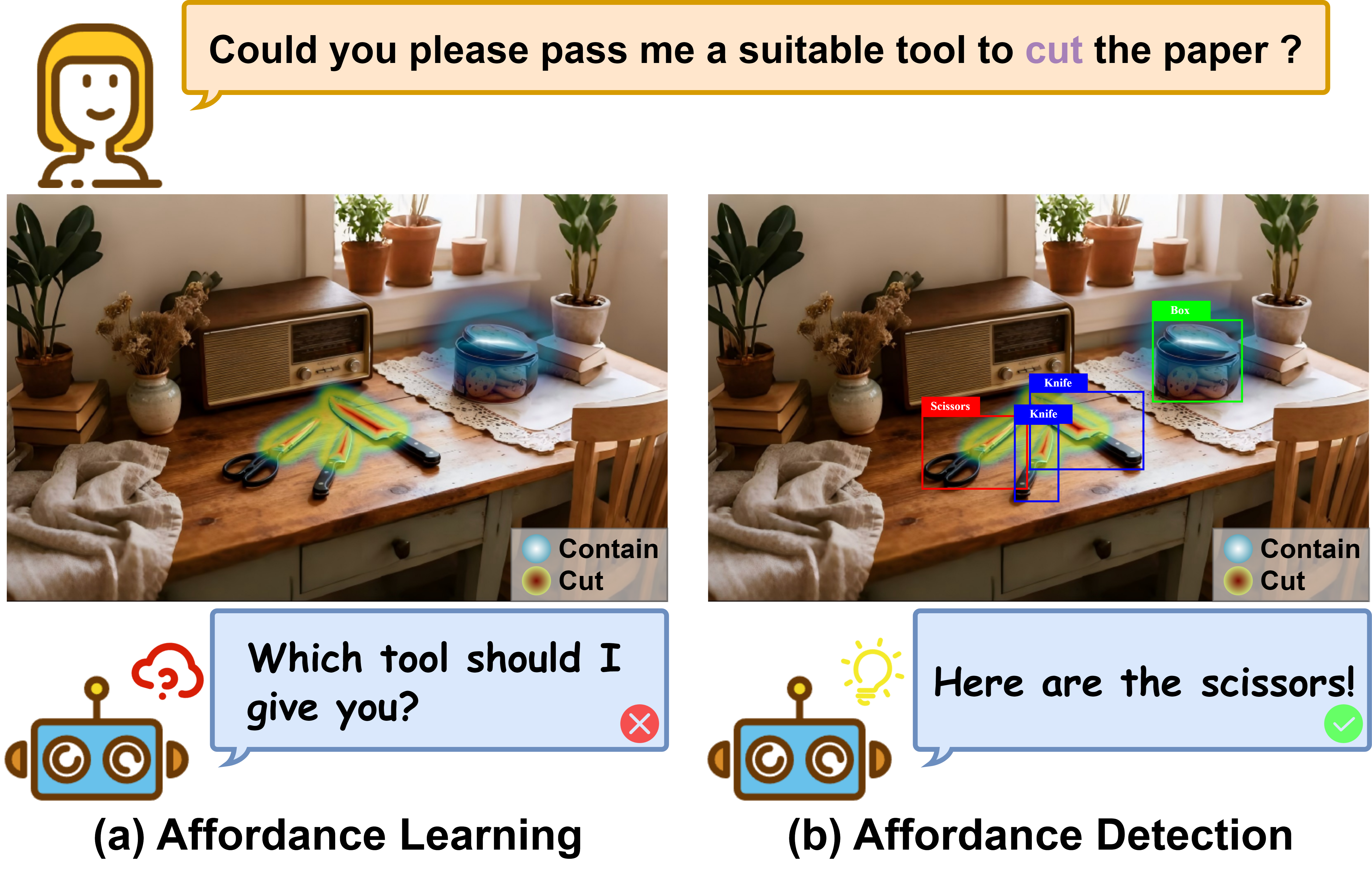}
	\caption{Comparison between Affordance Learning and Affordance Detection. The user requests a suitable tool to ``cut'' the paper. Robot A in \textbf{(a)} only identifies functional regions without recognizing object categories, making it unable to decide which tool to hand over. In contrast, Robot B in \textbf{(b)} integrates object categories, spatial locations, and affordances to accurately locate and identify the correct tool, enabling successful task completion.}
	\label{fig:difference}
	\vspace{-1em}
\end{figure}

Affordance detection~\cite{Do18, nguyen2016affordancecnn,nguyen2017affordancecrf,sawatzky2017weakly,luo2021oneshotaffordance,zhai2022oneshotaffordance} aims to perform object detection and affordance learning simultaneously, which is practical as it enables the interaction among ``what'', ``where'' and ``how''. In human-robot interaction (HRI)~\cite{Xiang2020SAPIEN,Li2025RVI,Wang2025RARV} and embodied AI~\cite{Lee2025DynScene,Savva2019Habitat,Deitke2020RoboTHOR,Wang2024EmbodiedScan}, a robot may encounter situations in which it stands before a cluttered table as a human asks, ``Could you please pass me a suitable tool to cut the paper?'' For the robot, affordance learning~\cite{chao2015semanticaffordance,fang2018demo2vec,li2024ooal,li2019putting,nagarajan2019hotspots}  merely enables it to understand  ``how'' objects can be used, which is insufficient. The robot also needs to understand ``what'' each object is and ``where'' it is located. As illustrated in Fig.~\ref{fig:difference}, both a knife and scissors can cut, but the knife is obviously not a sensible choice for cutting paper. However, this seemingly simple scenario reveals a fundamental challenge in affordance learning: it cannot be decoupled from object detection~\cite{myers2015toolaffordance, nguyen2016affordancecnn,nguyen2017affordancecrf,Do18,nagarajan2019hotspots,wang2025reasoningmamba}. A truly intelligent system must therefore integrate category, location, and affordance into a unified paradigm.

With the growing interest in HRI and embodied AI, affordance learning has developed rapidly in recent years. However, this progress remains limited in scope, as real-world scenarios require the “what–where–how” to be addressed jointly, while most previous methods~\cite{qian2024affordancellm,zhu2025lmaffordance3d,zhang2025iaao,wu2025afforddp,lu2025geal,shao2025great} still focus primarily on the “how” aspect. Such methods often neglect object category and spatial localization, even though these cues are essential for reliable affordance reasoning. In particular, recognizing a “knife” facilitates inferring affordances such as “cut” or “hold,” and spatial context further helps localize their interactive regions. These limitations highlight the need to move beyond affordance learning toward affordance detection, where category, location, and affordance are jointly estimated. Nevertheless, affordance detection remains largely underexplored, and existing methods~\cite{nguyen2017affordancecrf,Do18,luo2021oneshotaffordance,zhai2022oneshotaffordance,sawatzky2017weakly} typically treat object detection and affordance learning as independent tasks. This decoupling limits the synergy between object perception and functional learning, hindering unified and context-aware affordance reasoning.

To address these challenges, we propose YOLO Affordance (\textbf{YOLOA}), a real-time affordance detection model that unifies object detection and affordance learning tasks through a language-guided interaction and refinement mechanism. Inspired by recent advances in large language models (LLMs)~\cite{brown2020gpt3,touvron2023llama,chowdhery2023palm,achiam2023gpt4,bai2023qwen,abdin2025phi4}, we employ LLMs as a bridge to enable interaction between the object detection and affordance learning branches. Specifically, YOLOA is built upon a real-time detector~\cite{Khanam24} that comprises two parallel branches, and integrates an \textbf{LLM Adapter}~\cite{Hu22} to refine the preliminary predictions through dual-branch interaction. During training, the dual-branch framework outputs object and affordance preliminary predictions. Subsequently, the LLM Adapter leverages a frozen LLM to enhance these predictions through three language-guided refinements:
(1) Class priors, which provide semantic context for object classification;
(2) Box offsets, which refine target localization predictions; and
(3) Affordance gates, which constrain functional regions and guide regional affordance estimation. Moreover, the LLM Adapter can be removed after training to enable real-time inference, forming the lightweight variant \textbf{YOLOA-light}, which achieves up to 846.24 FPS while maintaining high accuracy, demonstrating strong potential for embodied AI deployment.

Our contributions are summarized as follows:
\begin{enumerate}[label=\textbullet]
	\item The proposed YOLOA integrates an affordance learning branch into a unified framework that jointly performs object detection and affordance learning in real-time, addressing the fundamental ``what–where–how" challenge in HRI and embodied AI.
	\item To the best of our knowledge, YOLOA is the first method to employ an LLM Adapter for interaction between the object detection and affordance learning branches, enablong mutual refinement and knowledge distillation from the LLM into both branches.
	\item We relabel two affordance datasets, ADG20K and IIT-AFF, to better reflect realistic embodied scenarios and evaluate YOLOA on the resulting benchmarks where it achieves state-of-the-art performance.
\end{enumerate}

\section{Related Work}

\subsection{Affordance Learning}

Affordance learning aims to associate visual appearance with object uses and potential actions.
Early studies~\cite{thermos2017deepaffordance, wang2017transferringobjects, fang2018demo2vec} construct affordance representations using auxiliary signals such as geometry, depth, or human demonstrations. 
Subsequent methods improve data efficiency through weakly supervised and cross-view learning paradigms~\cite{sawatzky2017weakly,luo2022affordancegrounding,li2024ooal}, while methods like LOCATE~\cite{li2023locate} further explicitly model object parts to support region-level affordance localization. 
R-Mamba~\cite{wang2025reasoningmamba}  utilizes higher-order region dependencies through a hypergraph-guided reasoning module.
Recent approaches combine visual, linguistic, and spatial cues to achieve more robust and generalizable affordance reasoning~\cite{qian2024affordancellm,zhu2025lmaffordance3d,zhang2025iaao,wu2025afforddp,lu2025geal,shao2025great}. 
However, these methods still fail to fully leverage category and localization information for affordance learning, limiting their ability to achieve more practical affordance reasoning.

\subsection{Affordance Detection}
Affordance detection plays a pivotal role in embodied AI by integrating object detection and affordance learning~\cite{Do18}. Early methods such as AffordanceCNN~\cite{nguyen2016affordancecnn} and AffordanceCRF~\cite{nguyen2017affordancecrf} leverage RGB-D inputs to predict affordance masks with refined spatial boundaries. Subsequent studies focus on improving data efficiency through weak supervision~\cite{sawatzky2017weakly}, one-shot affordance detection~\cite{luo2021oneshotaffordance}, and open-world generalization~\cite{zhai2022oneshotaffordance}.
More recently, researchers further integrate affordance reasoning into object-level perception, such as ATL~\cite{hou2021affordancetransfer} and CoTDet~\cite{tang2023cotdet}, which use affordance cues to enhance human–object interaction and object detection, respectively. 
Nevertheless, these approaches remain limited in achieving mutual enhancement between object detection and affordance learning, leaving their intrinsic complementarity largely underexplored.

\begin{figure*}[!t]
	\centering
	\includegraphics[width=\textwidth]{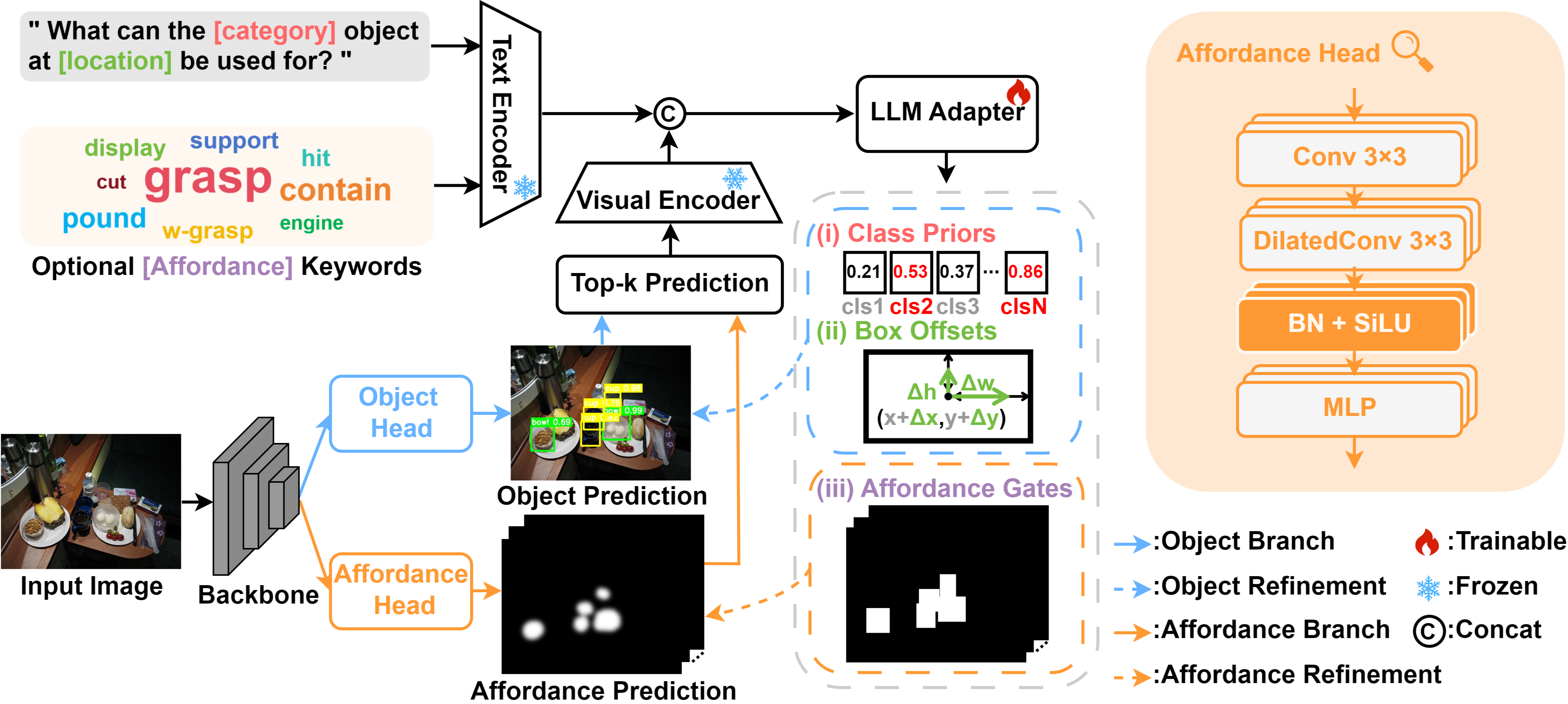}
	\caption{An overview of the proposed \textbf{YOLOA}.
		The backbone produces object and affordance predictions, which are integrated through a language-guided adapter. The LLM Adapter enhances both object detection and affordance learning branches through three semantic refinements, namely \textcolor[HTML]{FF6666}{class priors}, \textcolor[HTML]{75BD42}{box offsets}, and \textcolor[HTML]{A680B8}{affordance gates}.}
	\label{fig:overview}
	\vspace{-0.5em}
\end{figure*}

\subsection{LLM-Based Affordance Application}
Large language models (LLMs) demonstrate remarkable capabilities in reasoning~\cite{wei2022cot,zhou2023leasttomost}, common-sense understanding~\cite{sap2019atomic,bosselut2019comet}, and vision–language relation extraction~\cite{alayrac2022flamingo,Li23b,Liu23b}, which motivate their growing application in affordance reasoning. Recent studies, such as AffordanceLLM~\cite{qian2024affordancellm} for affordance grounding through vision–language models, SeqAfford~\cite{yu2025seqafford} for sequential 3D affordance reasoning, IAAO~\cite{zhang2025iaao} for interactive affordance learning of articulated objects, and LMAffordance3D~\cite{zhu2025lmaffordance3d} for instruction-guided 3D grounding, collectively demonstrate that LLM knowledge effectively guides affordance perception. Unlike these LLM-centric frameworks, our approach achieves lightweight integration by distilling LLM knowledge into the affordance head, enabling efficient real-time object–affordance detection.

\section{Method}

\textbf{Problem Formulation for Affordance Detection.} We consider an object–affordance labeled dataset, denoted as $\{X, C, B, A\}$. The $i$-th input sample consists of an input image $x_i \in X$, its corresponding category labels $c_i \in C$, bounding box coordinates $b_i \in B$,  and an affordance mask $a_i \in A$ that highlights functional regions of the objects.  The objective of affordance detection is to jointly learn two branches: an object detection branch $\mathcal{F}_{\text{det}}$, which accurately classifies and localizes multiple objects within the input image $x_i$; and an affordance learning branch $\mathcal{F}_{\text{aff}}$, which generates a dense affordance mask highlighting spatial regions that likely enable specific functional interactions.
\subsection{Overview}
Our proposed model YOLOA builds upon the well-known YOLOv11~\cite{Khanam24} architecture and incorporates an LLM Adapter that facilitates bidirectional interaction between the object detection branch~\cite{Li22} and the affordance learning branch. The overall pipeline is illustrated in Fig.~\ref{fig:overview}.

Given an input image $x_i$, the YOLOv11 backbone extracts multi-scale features and feeds them into two parallel branches: an object detection branch that predicts category logits and bounding boxes $(\hat{c}_i, \hat{b}_i)$, and an affordance learning branch designed to produce a dense affordance mask $\hat{a}_i$. The preliminary predictions from both branches are further integrated into a compact LLM Adapter module, which leverages a frozen large language model (LLM)~\cite{touvron2023llama} 
with LoRA-based fine-tuning~\cite{Hu22}. The LLM Adapter receives a set of structured visual–textual tokens, including top-$k$ predictions $(\hat{c}_{i}^{k}, \hat{b}_{i}^{k})$, their corresponding affordance responses $\hat{a}_{i}^{k}$, and task-specific queries describing potential object–affordance relations. This integration enables the LLM Adapter to reason jointly over visual cues and linguistic priors in a unified latent space.

Through joint reasoning within the language-guided latent space, the LLM Adapter produces a set of object and affordance refinements:
(1) Class priors $\overline{c}_{i}$ that modulate the classification logits of the object detection branch;
(2) Box offsets $\overline{b}_{i}$ that refine both the position and the scale of bounding boxes; and
(3) Affordance gates $\overline{a}_{i}$ that constrain functional regions and guide regional affordance estimation.
Together, these refinements are subsequently fed into the YOLO heads, enabling language-guided cues that jointly enhance both branches.

During training, the entire model is optimized end-to-end under a unified objective that combines detection, affordance, and language-guided refinement losses to ensure consistent supervision across visual and textual modalities. This joint optimization thereby promotes visual–linguistic alignment~\cite{Guo25} and enables the model to learn affordance detection through the cooperative interaction between object perception and functional learning~\cite{Do18}.

\subsection{Affordance Detection Framework}
This section details the proposed dual-branch framework that integrates object detection and affordance learning branches. Following the one-stage design of YOLOv11, we first employ a Darknet backbone to extract multi-scale features from input image $x_i$, formulated as:
\begin{equation}\label{eq:feature_extractor}
		p_{i}= \mathcal{\phi}(x_{i}),
\end{equation}
where $\mathcal{\phi}$ denotes the trainable Darknet backbone, and $p_i$ represents the extracted multiscale feature maps.

A dual-branch framework is adopted to produce both object and affordance preliminary predictions from the multi-scale feature maps $p_i$, formulated as:
\begin{equation}
	\label{eq:dual-output}
	\begin{aligned}
		(\hat{c}_i,\hat{b}_i) &= \mathcal{F}_{\text{det}}(p_i),\\
		\hat{a}_i &= \mathcal{F}_{\text{aff}}(p_i).
	\end{aligned}
\end{equation}
\noindent
as defined above, $\mathcal{F}_{\text{det}}$ and $\mathcal{F}_{\text{aff}}$ denote the object detection and affordance learning branches, respectively.
Specifically, $\mathcal{F}_{\text{aff}}$ applies multi-scale visual features through a lightweight
convolutional block composed of a standard $3{\times}3$ convolution, a dilated convolution~\cite{Yu16} ($\mathrm{DConv}$), normalization, activation, and a multilayer perceptron:
\begin{equation}
	\label{eq:aff-branch}
	\mathcal{F}_{\text{aff}}(p_i)
	=
	\mathrm{MLP}\!\Big(
	\mathrm{SiLU}\!\big(
	\mathrm{BN}\!\big(
	\mathrm{DConv}\!\big(\mathrm{Conv}(p_i)
	\big)\big)\big)\Big).
\end{equation}

Meanwhile, the object detection branch $\mathcal{F}_{\text{det}}$ is identical to that in YOLOv11, with the detection loss is defined as:
\begin{equation}
	\label{eq:detloss}
	\mathcal{L}_{\text{det}} =  
	\mathrm{BCE}(\hat{c}_i, c_i) +
	\mathrm{IOU}(\hat{b}_i, b_i) +
	\mathrm{DFL}(\hat{b}_i, b_i),
\end{equation}
where $\mathrm{BCE}$ denotes the binary cross-entropy loss~\cite{rumelhart1986backprop}, while $\mathrm{IOU}$ and $\mathrm{DFL}$ represent the IoU loss~\cite{yu2016unitbox} and the distribution focal loss~\cite{li2020gfl}, respectively.

The affordance loss $\mathcal{L}_{\text{aff}}$ is defined by  binary cross-entropy loss for $\mathcal{F}_{\text{aff}}$, designed to capture fine-grained functional cues while preserving the real-time constraint, as:
\begin{equation}
	\label{eq:affloss}
	\mathcal{L}_{\text{aff}} =\mathrm{BCE}(\hat{a}_i, a_i).
\end{equation}

After obtaining the preliminary predictions, we introduce the LLM Adapter to refine them in the next section. 

\subsection{Large Language Model Adapter}

To enable effective interaction between the two parallel branches, we introduce the LLM Adapter, which jointly processes the preliminary visual predictions $(\hat{c}_i,\hat{b}_i)$ and $\hat{a}_i$ with textual prompts. The LLM Adapter leverages a frozen LLM~\cite{touvron2023llama} with LoRA-based fine-tuning~\cite{Hu22} to perform cross-modal reasoning between object and affordance predictions. Serving as a bridge between the two branches, it facilitates bidirectional information and semantic alignment through joint visual–textual embedding~\cite{Li23b,Liu23b}.

Based on the object and affordance preliminary predictions $(\hat{c}_i,\hat{b}_i), \hat{a}_i$, two complementary embeddings are constructed for the LLM.\footnote{Following standard multimodal warm-up strategies~\cite{chen2020uniter,zhu2021deformable}, ground-truth labels $({c}_i,{b}_i)$, ${a}_i$  are used in early training to stabilize optimization and avoid noisy predictions.} 
1) Visual embedding.
For top-$k$ predictions $(\hat{c}_{i}^{k}, \hat{b}_{i}^{k})$, the corresponding region of the input image $x_i$ and predicted affordance mask $\hat{a}_{i}$ is cropped according to the predicted bounding box $\hat{b}_{i}^{k}$ and masked to retain only the target area. Each cropped region is then pooled with ROIAlign~\cite{he2017maskrcnn} to obtain a fixed-size representation before being linearly projected into the shared embedding space, yielding the visual embedding $E_{\text{vis}}$.
2) Text embedding. The top-$k$ predictions are further converted into location-aware, question-style textual prompts that describe object categories, (e.g., “What can the [category] object at [location] be used for?”, where [category] and [location] correspond to $\hat{c}_{i}^{k}$ and $\hat{b}_{i}^{k}$, respectively). These prompts and potential affordance keywords are then encoded by the LLM into the textual embeddings $E_{\text{text}}$.

The visual and text embeddings are then concatenated and fed into the LLM~\cite{chen2020uniter,zhang2021vinvl}:
\begin{equation}
\{h_{i}^{t}\}_{t=1}^{T}= \mathrm{LLM}(\mathrm{Concat}(E_{\text{vis}}, E_{\text{text}})),
\end{equation}
where ${h_{i}^{t}}$ denotes the hidden representations of the fused sequence, and $T$ is the total number of tokens after concatenation.
The hidden vector corresponding to the last textual token $h_{i}^{T}$ 
is fed into three lightweight components of the LLM Adapter, each responsible for a distinct refinement:
\begin{equation}
	\begin{aligned}
		\overline{c}_{i}^{k} &= \mathcal{F}_{\text{adapter}}^{\text{cls}}(h_{i}^{T}),\\
		\overline{b}_{i}^{k} &=
		 \mathcal{F}_{\text{adapter}}^{\text{box}}(h_{i}^{T}),\\
		\overline{a}_i &= \mathcal{F}_{\text{adapter}}^{\text{aff}}(h_{i}^{T}, \hat{b}_{i}^{k}).
	\end{aligned}
\end{equation}
where $\overline{c}_{i}^{k}$ and $\overline{b}_{i}^{k}$ denote the class priors and box offsets that refine the corresponding top-$k$ object predictions, respectively.
The affordance prediction is spatially expanded and masked by the detected bounding boxes $\hat{b}_{i}^{k}$ to form spatial affordance gates $\overline{a}_i$, which suppresses activations outside the object regions. 

The language-guided refinements produced by the LLM Adapter are propagated back into the object detection and affordance learning branches, forming a closed-loop interaction that updates predictions with semantic refinement:
\begin{equation}
	\begin{aligned}
		(\hat{c}_{i}^{k}, \hat{b}_{i}^{k}) &\leftarrow \big(\hat{c}_{i}^{k} + \alpha \cdot \overline{c}_{i}^{k}, \hat{b}_{i}^{k} + \beta\cdot\overline{b}_{i}^{k}\big),\\
		\hat{a}_{i} &\leftarrow \hat{a}_{i} + \gamma \cdot \operatorname{logit}(\overline{a}_{i}).
	\end{aligned}
	\label{eq:refine-all}
\end{equation}

We train this closed-loop interaction by supervising each LLM Adapter component with classification, localization, and affordance losses:
\begin{equation}
	\begin{aligned}
		\mathcal{L}_{\text{cls-priors}} 
		&= \mathrm{BCE}(\hat{c}_{i}^{k}, c_i),\\
		\mathcal{L}_{\text{box-offsets}} 
		&= {\mathrm{Smooth}\text{-}\ell_1}(\hat{b}_{i}^{k}, b_i),\\
		\mathcal{L}_{\text{aff-gates}} 
		&= \mathrm{BCE}(\hat{a}_{i}^{k}, a_i). 
    \end{aligned}
\end{equation}
\begin{table*}[t]
	\caption{Statistics of existing image-based affordance datasets and the relabeled datasets (ADG-Det and IIT-Heat) used in our experiments. \textbf{\#Aff.}, \textbf{\#Obj.}, and \textbf{\#Img.} indicate the numbers of affordance categories, object categories, and images, respectively.}
	\label{tab:dataset_comparison}
	\centering
	\makebox[\textwidth][c]{
		\resizebox{\textwidth}{!}{ 
			\tiny               
			\begin{tabular}{l@{\hspace{5pt}}cccccc}
				\toprule
				\textbf{Dataset} & \textbf{Publication} & \textbf{Affordance Format} &
				\textbf{Object Boxes} & \textbf{\#Aff.} & \textbf{\#Obj.} & \textbf{\#Img.} \\
				\midrule
				UMD~\cite{myers2015toolaffordance} & ICRA 2015 & Segmentation & \ding{55} & 7 & 17 & 30,000 \\
				IIT-AFF~\cite{nguyen2017affordancecrf} & IROS 2017 & Segmentation & \ding{51} & 9 & 10 & 8,835 \\
				OPRA~\cite{fang2018demo2vec} & CVPR 2018 & Heatmap & \ding{55} & 7 & -- & 5,500 \\
				ADE-AFF~\cite{chuang2018actproperly} & CVPR 2018 & Segmentation & \ding{55} & 7 & 150 & 10,000 \\
				PADv2~\cite{zhai2022oneshotaffordance} & IJCV 2022 & Segmentation & \ding{55} & 39 & 103 & 30,000 \\
				ADG20K~\cite{luo2022affordancegrounding} & CVPR 2022 & Heatmap & \ding{55} & 36 & 50 & 20,000 \\
				\midrule
				\textbf{ADG-Det (relabeled)} &   & Heatmap & \ding{51} & 36 & 50 & 2,000 \\
				\textbf{IIT-Heat (relabeled)} &   & Heatmap & \ding{51} & 9 & 10 & 8,835 \\
				\bottomrule
			\end{tabular}
		}
	}
	\vspace{-0.5em}
\end{table*}

The overall LLM Adapter loss aggregates all semantic supervision components. Importantly, during joint optimization, both branches remain frozen while the LLM Adapter learns to align their representations through semantic supervision:
\begin{equation}
	\mathcal{L}_{\text{adapter}} =
	\lambda_1 \mathcal{L}_{\text{cls-priors}} +
	\lambda_2 \mathcal{L}_{\text{box-offsets}} +
	\lambda_3 \mathcal{L}_{\text{aff-gates}}.
\end{equation}

The proposed supervision encourages the LLM Adapter to produce semantically meaningful refinements aligned with both object-level and affordance-level cues, thereby promoting mutual enhancement between visual perception and linguistic reasoning for affordance detection.

The overall optimization objective combines detection, affordance, and language-guided refinement losses as:
\begin{equation}
	\mathcal{L}_{\text{total}} = 
	\mathcal{L}_{\text{det}} + \mathcal{L}_{\text{aff}} + \mathcal{L}_{\text{adapter}}.
\end{equation}

This unified loss ensures consistent optimization across visual and linguistic modalities, enabling the model to jointly learn object perception and affordance reasoning in an end-to-end manner.

\subsection{Lightweight Inference YOLOA}
Although YOLOA benefits from the prediction refinement of the LLM Adapter during training, such language-guided reasoning is not necessary during inference. To achieve real-time efficiency, we introduce a lightweight variant, denoted as YOLOA-light, which disables the LLM Adapter during inference and directly performs prediction through the dual-branch framework. 
This design follows the decoupled paradigm similar to prior multimodal detectors~\cite{Li23b,Liu23b}, where semantic reasoning is leveraged only during training to enhance feature alignment, but is omitted during inference for efficiency.

In YOLOA-light, the object detection branch $\mathcal{F}_{\text{det}}$ outputs object categories and bounding boxes, while the affordance learning branch $\mathcal{F}_{\text{aff}}$ produces affordance masks without semantic refinement. Since the LLM Adapter remains frozen, the learned visual–linguistic alignment is implicitly retained in the backbone, allowing the lightweight variant to maintain semantic consistency while operating with lower inference time cost than the full configuration. As demonstrated in Tab.~\ref{tab:sota_comparison}, YOLOA-light achieves a favorable trade-off between accuracy and real-time efficiency.

\section{Experiments}
\subsection{Dataset Preparation}
Most affordance datasets lack object-level annotations and rely on costly segmentation masks. To fit the affordance detection paradigm, we relabel two representative datasets, ADG20K and IIT-AFF. Table~\ref{tab:dataset_comparison} compares existing datasets with our detection-oriented relabeled datasets.

\textbf{ADG-Det.} The ADG20K dataset~\cite{luo2022affordancegrounding} offers cross-view affordance annotations and is widely adopted in the affordance learning task. To ensure category diversity, we manually annotated 1,000 images each from exocentric and egocentric views with object categories and bounding boxes.
The constructed ADG-Det dataset contains 2,000 images, 36 affordance categories, and 50 object classes, serving as a comprehensive benchmark for affordance detection.

\begin{figure*}[t]
	\centering
	\includegraphics[width=\textwidth]{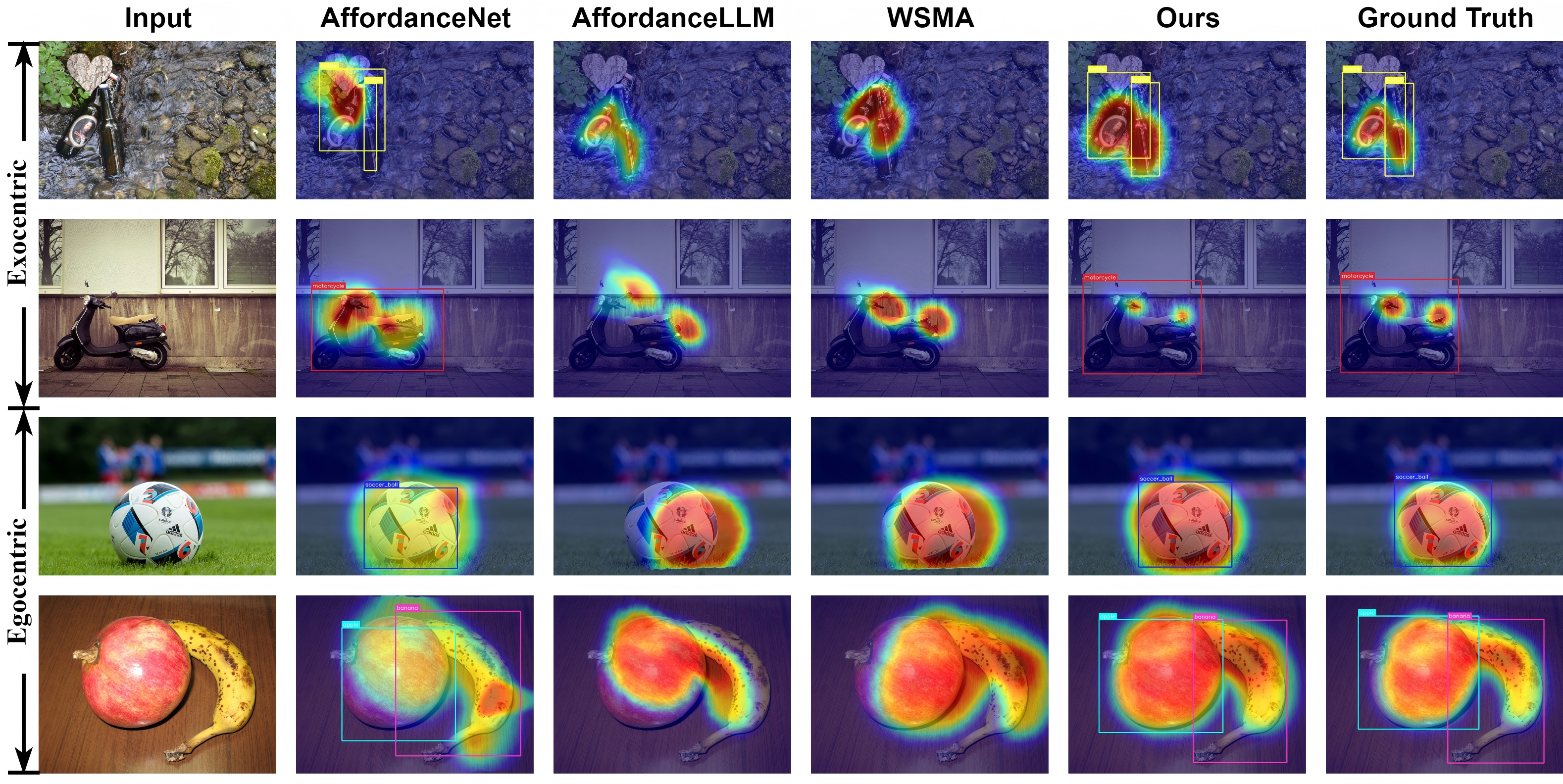}
	\caption{Qualitative visualization on the ADG-Det dataset. Each column shows the predictions of a different method (including the input image and ground truth) for a given affordance category, visualized under exocentric (top) and egocentric (bottom) views.}
	\label{fig:vis_ADG}
	\vspace{-1em}
\end{figure*}

\textbf{IIT-Heat.} The IIT-AFF dataset~\cite{nguyen2017affordancecrf} was captured in cluttered tabletop scenes with segmentation masks and bounding boxes for realistic HRI scenarios. However, dense segmentation labeling is costly and hard to scale~\cite{myers2015toolaffordance,sawatzky2017weakly,Do18}, while recent studies~\cite{li2024ooal,hou2021affordancetransfer,sawatzky2017weakly} indicate that coarse or keypoint-level supervision provides comparable performance for affordance learning. To reduce annotation cost while preserving spatial cues, we follow Demo2Vec’s keypoint strategy~\cite{fang2018demo2vec} to convert affordance annotations, originally represented as segmentation masks, into heatmaps.
The resulting IIT-Heat dataset includes 8,835 images, 9 affordance categories, and 10 object classes, serving as an efficient benchmark for weakly supervised multi-object, multi-affordance detection.

\subsection{Benchmark Setup}

\textbf{Evaluation Metrics.} We evaluate object and affordance prediction using two sets of metrics. For object prediction, we adopt mean Average Precision (\textbf{mAP}) and Average Recall (\textbf{AR}), which are commonly adopted in object detection~\cite{girshick2015fast,ren2016faster,lin2017fpn}.  Here, mAP reflects the overall detection accuracy, while AR measures the average recall of detected objects. Following previous work~\cite{li2023locate,luo2022affordancegrounding,li2024ooal,qian2024affordancellm}, the quality of affordance prediction is evaluated using heatmap-based metrics: Kullback–Leibler Divergence (\textbf{KLD}), Similarity (\textbf{SIM}), and Normalized Scanpath Saliency (\textbf{NSS}). Together, these metrics comprehensively assess the accuracy and spatial consistency of affordance detection.

\noindent\textbf{Implementation Details.} YOLOA is built upon YOLOv11 as the base detector and incorporates LLaMA~\cite{touvron2023llama} through an LLM Adapter. All experiments are implemented in PyTorch and trained on four NVIDIA A100 GPUs using the SGD optimizer (momentum 0.9, weight decay $5{\times}10^{-4}$, batch size 64).
The training runs for 800 epochs with 10 warm-up epochs, and the learning rate is scheduled by cosine annealing from $1{\times}10^{-8}$ to $2{\times}10^{-3}$.

\begin{table*}[!t]
	\caption{Unified comparison on the ADG-Det and IIT-Heat datasets. \textbf{Obj.}/\textbf{Aff.} indicate whether a method performs object detection / affordance learning. Best and second-best are in \textbf{bold} and \underline{underlined}, respectively. ($\uparrow/\downarrow$ denote higher/lower is better.)}
	\label{tab:sota_comparison}
	\centering
	\resizebox{\textwidth}{!}{\setlength{\heavyrulewidth}{1.2pt}
		\begin{tabular}{l@{\hspace{3pt}}c@{\hspace{3pt}}cc@{\hspace{5pt}}c@{\hspace{5pt}}c@{\hspace{5pt}}c@{\hspace{5pt}}c@{\hspace{4pt}}cc@{\hspace{4pt}}c@{\hspace{5pt}}c@{\hspace{5pt}}c@{\hspace{5pt}}c@{\hspace{5pt}}c}
			\toprule
			\multirow[c]{2}{*}{\vspace{-5pt}\textbf{Method}} &
			\multirow[c]{2}{*}{\vspace{-5pt}\textbf{Obj.}} &
			\multirow[c]{2}{*}{\vspace{-5pt}\textbf{Aff.}} &
			\multicolumn{6}{c}{\textbf{ADG-Det}} &
			\multicolumn{6}{c}{\textbf{IIT-Heat}} \\
			\cmidrule(lr){4-9} \cmidrule(lr){10-15}
			& & & mAP$\uparrow$ & AR$\uparrow$ & KLD$\downarrow$ & SIM$\uparrow$ & NSS$\uparrow$ & \textbf{FPS}$\uparrow$
			& mAP$\uparrow$ & AR$\uparrow$ & KLD$\downarrow$ & SIM$\uparrow$ & NSS$\uparrow$ & \textbf{FPS}$\uparrow$ \\
			\midrule
			AffordanceNet~\cite{Do18} & \ding{51} & \ding{51}
			& 37.2 & 30.1 & 2.788 & 0.162 & 0.579 & 6.41
			& 44.6 & 52.7 & 4.083 & 0.138 & 0.842 & 10.73 \\
			
			Cross-View-AG~\cite{luo2022affordancegrounding} & \ding{55} & \ding{51}
			& -- & -- & 1.829 & 0.241 & 0.928 & 41.82
			& -- & -- & 3.856 & 0.097 & 0.834 & 43.58 \\
			
			LOCATE~\cite{li2023locate} & \ding{55} & \ding{51}
			& -- & -- & 1.773 & 0.309 & 1.218 & \underline{77.56}
			& -- & -- & 3.328 & 0.146 & 1.425 & 86.94 \\
			
			CoTDet~\cite{tang2023cotdet} & \ding{51} & \ding{55}
			& 48.3 & 46.6 & -- & -- & -- & 20.84
			& 59.6 & 67.3 & -- & -- & -- & 32.37 \\
			
			AffordanceLLM~\cite{qian2024affordancellm} & \ding{55} & \ding{51}
			& -- & -- & 1.731 & 0.281 & 1.241 & 0.93
			& -- & -- & 2.831 & 0.152 & 1.387 & 1.38 \\

			WSMA~\cite{xu2024wsmag} & \ding{55} & \ding{51}
			& -- & -- & \underline{1.637} & \underline{0.327} & 1.267 & 61.74
			& -- & -- & \underline{2.674} & 0.163 & 1.528 & 67.31 \\
			
			\textbf{YOLOA (Ours)} & \ding{51} & \ding{51}
			& \textbf{52.8} & \textbf{49.7} & \textbf{1.528} & \textbf{0.341} & \textbf{1.346} & 73.76
			& \textbf{73.1} & \textbf{74.8} & \textbf{2.553} & \textbf{0.187} & \textbf{1.838} & \underline{89.77} \\
			
			\textbf{YOLOA-light (Ours)} & \ding{51} & \ding{51}
			& \underline{51.7} & \underline{47.8} & 1.659 & 0.312 & \underline{1.311} & \textbf{470.58}
			& \underline{72.6} & \underline{73.6} & 2.681 & \underline{0.171} & \underline{1.531} & \textbf{846.24} \\
			\bottomrule
		\end{tabular}
	}
    \vspace{-1em}
\end{table*}

\subsection{Performance Comparison}

To provide a comprehensive evaluation, we further compare YOLOA with recent state-of-the-art methods on both the ADG-Det and IIT-Heat datasets.

Table~\ref{tab:sota_comparison} presents the comparison with state-of-the-art affordance detection and learning methods on the ADG-Det and IIT-Heat datasets. 
Compared with existing approaches, our proposed YOLOA consistently achieves the best performance on both datasets across all metrics. On the ADG-Det dataset, YOLOA achieves the highest detection accuracy with an mAP of \textbf{52.8} and AR of \textbf{49.7}, surpassing CoTDet~\cite{tang2023cotdet} by 4.5 and 3.1 points, respectively. Meanwhile, it yields the lowest KLD (\textbf{1.528}) as well as the highest SIM (\textbf{0.341}) and NSS (\textbf{1.346}), indicating superior spatial consistency and attention alignment between predicted and ground-truth affordance masks. In contrast, prior affordance-focused methods such as AffordanceLLM~\cite{qian2024affordancellm} and WSMA~\cite{xu2024wsmag} exhibit limited localization capability, as they lack an explicit object detection branch, leading to less precise estimation of functional boundaries. On the IIT-Heat dataset, YOLOA further improves generalization to realistic scenes, achieving an mAP of \textbf{73.1} and AR of \textbf{74.8}, outperforming all previous methods by a substantial margin. It also attains the best results in heatmap-based metrics (KLD = \textbf{2.553}, SIM = \textbf{0.187}, NSS = \textbf{1.838}), demonstrating that the language-guided affordance reasoning effectively enhances both inter-class discriminability and region precision. Moreover, the superior results on both ADG-Det and IIT-Heat datasets demonstrate YOLOA’s strong generalization ability across diverse settings, including cross-view and multi-object affordance scenarios.

In addition, the lightweight variant YOLOA-light maintains comparable accuracy to the full configuration (51.7 / 72.6 mAP on ADG-Det / IIT-Heat) while significantly reducing computational complexity and retaining real-time performance. These results clearly indicate that the proposed language-guided interaction mechanism can be efficiently scaled down without substantial performance degradation. Notably, YOLOA-light also attains the second-best accuracy across most metrics and reaches an impressive inference performance (up to \textbf{846.24} FPS), highlighting its strong potential for real-time deployment.

\begin{table}[!t]
	\caption{Ablation study of YOLOA on the ADG-Det dataset, evaluating the contribution of each module.}
	\label{tab:ablation_llm_paths}
	\centering
	\resizebox{\columnwidth}{!}{\setlength{\heavyrulewidth}{1.2pt}
		\begin{tabular}{lc@{\hspace{5pt}}c@{\hspace{5pt}}c@{\hspace{5pt}}c@{\hspace{5pt}}c}
			\toprule
			\multirow[c]{2}{*}{\vspace{-5pt}\textbf{Setting}} &
			\multicolumn{5}{c}{\textbf{Metrics}} \\
			\cmidrule(lr){2-6}
			& \textbf{mAP}$\uparrow$ & \textbf{AR}$\uparrow$ & \textbf{KLD}$\downarrow$ & \textbf{SIM}$\uparrow$ & \textbf{NSS}$\uparrow$ \\
			\midrule
			w/o LLM Adapter &49.6 & 43.7 &1.812& 0.293 & 1.176  \\
			w/o Obj. Branch &--&--&1.739&0.315&1.244\\
			w/o Aff. Branch&50.2&46.3&--&--&--  \\
			\textbf{Full (Ours)}   & \textbf{52.8} & \textbf{49.7} & \textbf{1.528} & \textbf{0.341} & \textbf{1.346} \\
			\bottomrule
	\end{tabular}}
    \vspace{-1.5em}
\end{table}

\subsection{Qualitative Results}

\noindent\textbf{ADG-Det.} As illustrated in Fig.~\ref{fig:vis_ADG}, YOLOA exhibits clear and focused activation on task-relevant regions under both exocentric and egocentric views. Although AffordanceNet demonstrates the ability to jointly detect objects and corresponding functional regions, its performance in both tasks remains limited. Meanwhile, AffordanceLLM and WSMA tend to produce affordance masks that are improperly scaled or spatially misaligned with the target regions. In contrast, our method achieves well-constrained affordance localization by leveraging instance-level positional guidance from the object detection branch.

\begin{figure*}[t]
	\centering
	\includegraphics[width=\textwidth]{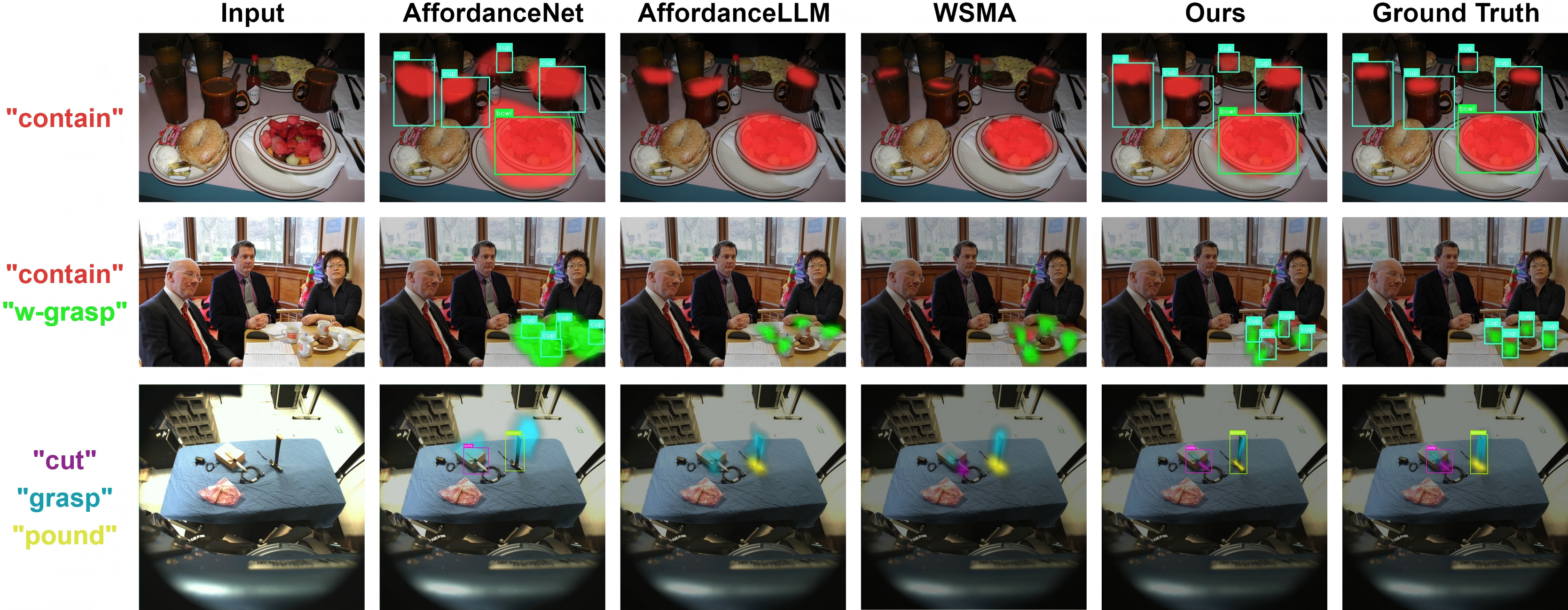}
	\caption{Qualitative visualization on the IIT-Heat dataset. The affordance categories for each image are listed on the left, with label colors matching those in the corresponding affordance masks on the right. (Each image may include multiple objects and affordances.)}
	\label{fig:vis_IIT}
	\vspace{-0.5em}
\end{figure*}

\noindent\textbf{IIT-Heat.} As shown in Fig.~\ref{fig:vis_IIT}, for images containing a single affordance category, previous methods correctly identify the corresponding category, although their precision in affordance localization varies. However, when multiple affordance categories coexist within the same image, both AffordanceNet and AffordanceLLM fail to detect all functional labels associated with each object. Unlike these methods, YOLOA accurately recognizes all affordance categories without omission by exploiting the object–affordance associations established through the LLM Adapter.

\begin{figure}[!t]
	\centering
	\includegraphics[width=\columnwidth]{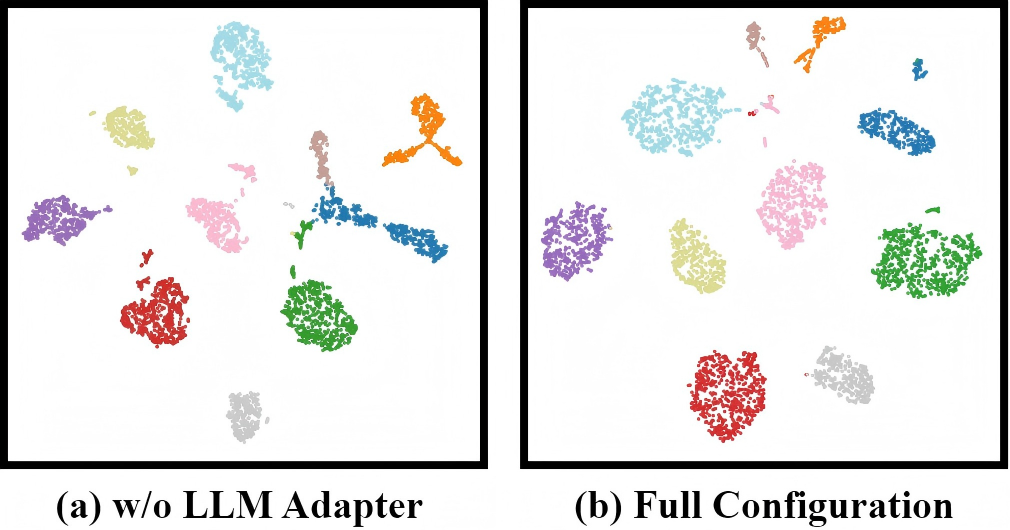}
	\caption{The t-SNE visualization on the IIT-Heat dataset, comparing (a) the model without the LLM Adapter and (b) the full configuration to illustrate differences in feature distribution.}
	\label{fig:t-SNE}
	\vspace{-1em}
\end{figure}

\subsection{Ablation Study}

\textbf{Ablation study of YOLOA.} To analyze the contribution of the proposed LLM Adapter to YOLOA, we conduct an ablation study summarized in Tab.~\ref{tab:ablation_llm_paths}.
When the LLM Adapter is removed, the performance on both branches degrades notably, highlighting that dual-branch interaction is essential for synergistic learning. In contrast, removing either the object detection or affordance learning branch (w/o Obj./Aff. Branch) still benefits from LLM guidance but remains clearly inferior to the full configuration, indicating that both branches are indispensable for dual-branch enhancement. We also visualize the feature distributions on the IIT-Heat dataset using t-SNE. 
As shown in Fig.~\ref{fig:t-SNE}, features with the LLM Adapter exhibit more compact and separable clusters, suggesting a clearer semantic structure in the embedding space. Additional visual comparisons between the full configuration and the model without the LLM Adapter are provided in the supplementary material.

\noindent\textbf{Ablation study of LLM Adapter.} We further analyze the LLM Adapter components responsible for generating class priors (CLS.), box offsets (BOX.), and affordance gates (AFF.). As presented in Tab.~\ref{tab:adapter_ablation}, activating any single component improves performance over the model without the LLM Adapter, indicating that each component contributes complementary cues for classification, localization, and affordance refinement. Among them, the component responsible for affordance gates provides the largest improvement by constraining spatial activations within object regions. When all three components are jointly enabled, the model achieves the best results, confirming the synergistic benefit of joint refinement under LLM guidance.

\begin{table}[!t] 
	\caption{
	Ablation  study of the three components within the LLM Adapter on the ADG-Det dataset. \textbf{CLS.}, \textbf{BOX.}, and \textbf{AFF.} denote the components of the LLM Adapter responsible for generating the class priors, box offsets, and affordance gates, respectively.
	}
	\label{tab:adapter_ablation}
	\centering
	\resizebox{\columnwidth}{!}{\setlength{\heavyrulewidth}{1.2pt}
		\begin{tabular}{l@{\hspace{6pt}}c@{\hspace{6pt}}cc@{\hspace{6pt}}c@{\hspace{5pt}}c@{\hspace{5pt}}c@{\hspace{5pt}}c}
			\toprule
			\multirow[c]{2}{*}{\vspace{-5pt}\textbf{CLS.}} &
			\multirow[c]{2}{*}{\vspace{-5pt}\textbf{BOX.}} &
			\multirow[c]{2}{*}{\vspace{-5pt}\textbf{AFF.}} &
			\multicolumn{5}{c}{\textbf{Metrics}} \\		
			\cmidrule(lr){4-8}
			& & & \textbf{mAP}$\uparrow$ & \textbf{AR}$\uparrow$ & \textbf{KLD}$\downarrow$ & \textbf{SIM}$\uparrow$ & \textbf{NSS}$\uparrow$ \\
			\midrule
			&  &  & 49.6 & 43.7 & 1.812 & 0.293 & 1.176 \\ 
			\ding{51} &  &  & 50.8 & 45.2 & 1.796 & 0.297 & 1.188 \\ 
			& \ding{51} &  & 51.7 & 45.5 & 1.808 & 0.284 & 1.163 \\ 
			&  & \ding{51} & 49.3 & 42.3 & 1.562 & 0.338 & 1.274 \\ 
			\ding{51} & \ding{51} &  & 52.0 & 48.9 & 1.881 & 0.280 & 1.228 \\ 
			\ding{51} &  & \ding{51} & 50.5 & 46.1 & 1.591 & 0.347 & 1.309 \\ 
			& \ding{51} & \ding{51} & 51.1 & 47.7 & 1.676 & \textbf{0.354} & 1.319 \\ 
			\ding{51} & \ding{51} & \ding{51} & \textbf{52.8} & \textbf{49.7} & \textbf{1.528} & 0.341 & \textbf{1.346} \\ 
			\bottomrule
	\end{tabular}}
	\vspace{-1.5em}
\end{table}

\begin{figure}[!t]
	\centering
	\includegraphics[width=\columnwidth]{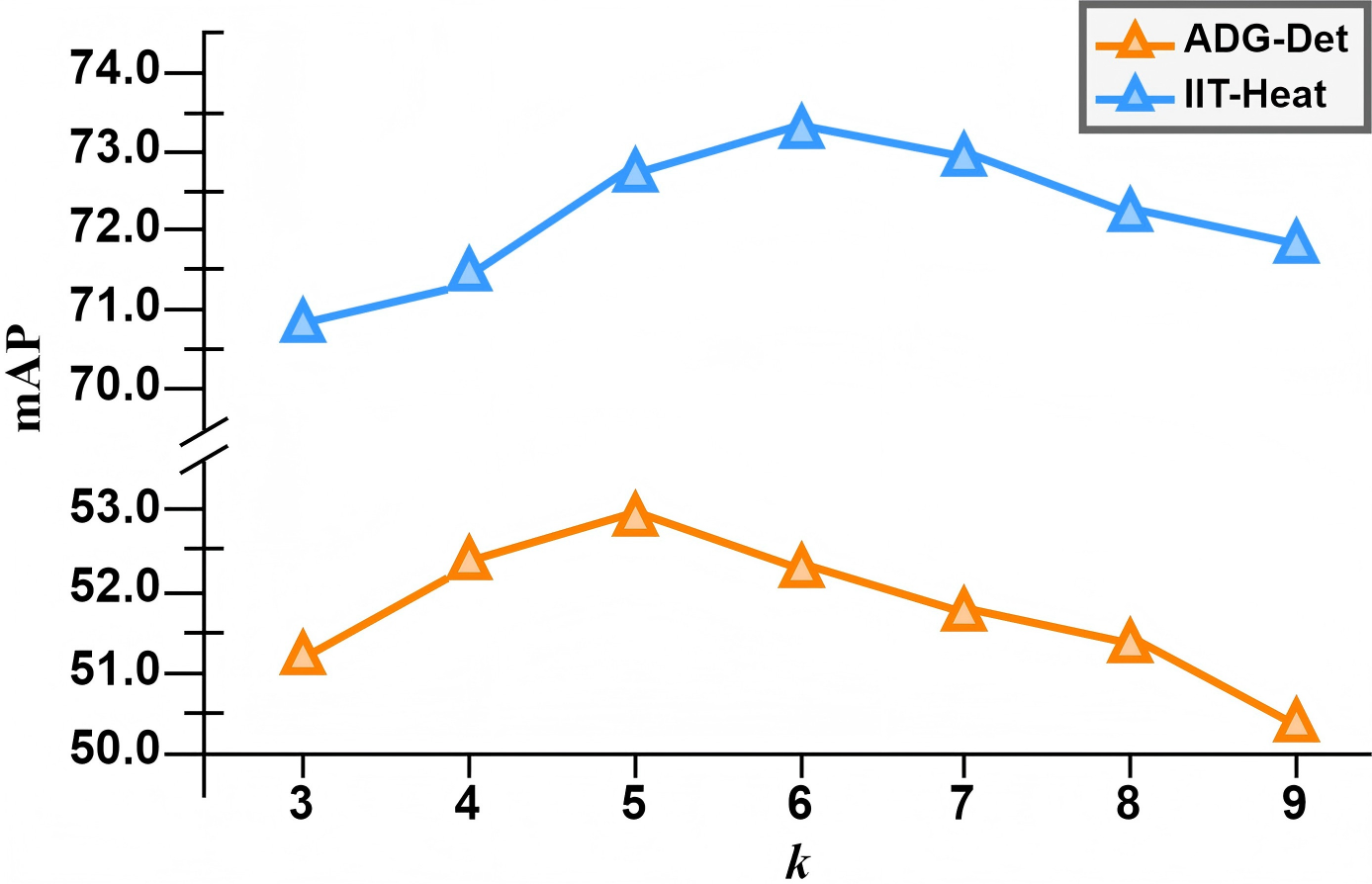}
	\caption{Ablation study on the number of top-$k$ predictions fed into the LLM Adapter on the ADG-Det and IIT-Heat datasets.}
	\label{fig:topk}
	\vspace{-1.5em}
\end{figure}

\noindent\textbf{Ablation study of Top-k Prediction.} We finally analyze how the number of top-$k$ predictions fed into the LLM Adapter affects overall performance. As shown in Fig.~\ref{fig:topk}, increasing $k$ initially improves the detection accuracy on both ADG-Det and IIT-Heat datasets by providing more informative contextual cues that enhance the LLM Adapter’s reasoning capability. However, when $k$ exceeds the average number of objects per image, low-confidence proposals introduce semantic noise and lead to a degradation in performance. The results show that a moderate value of $k$ (5 for ADG-Det, 6 for IIT-Heat) achieves the best trade-off between contextual completeness and noise robustness, yielding the highest detection accuracy across both datasets.

\section{Conclusion}

In this work, we propose YOLOA, a real-time affordance detection model that jointly handles object detection and affordance learning tasks. Unlike conventional affordance learning approaches that focus solely on ``how" objects can be used, YOLOA unifies the ``what–where–how" relationship through a semantic interaction between the object detection and affordance learning branches. Such a  design overcomes the limitations of existing decoupled methods. We believe this work lays the foundation for real-time affordance detection in HRI and embodied AI, bridging perception and interaction through language-guided reasoning.
{
    \small
    \bibliographystyle{ieeenat_fullname}
    \bibliography{main}
}

\clearpage          
\appendix           

\setcounter{table}{0}
\setcounter{figure}{0}
\setcounter{equation}{0}


\renewcommand{\thetable}{S\arabic{table}}
\renewcommand{\thefigure}{S\arabic{figure}}
\renewcommand{\theequation}{S\arabic{equation}}

\twocolumn[{%
	\centering
	{\Large\bfseries YOLOA: Real-Time Affordance Detection via LLM Adapter\\[3em]
		\large\bfseries Supplementary Materials\par}
	\vspace{4em}
}]

\section*{Supplementary Overview}

\begin{figure*}[!t]
	\centering
	\includegraphics[width=\linewidth]{Figure/wo_IIT.png}
	\caption{\textbf{Visualization of the LLM Adapter effects on the IIT-Heat dataset.} The comparison between YOLOA without and with the LLM Adapter shows that the adapter consistently improves both object detection accuracy and affordance learning quality, in comparison with the ground-truth annotations. The affordance categories for each image are listed on the left, with label colors matching those in the corresponding affordance masks on the right.}
	\label{fig:vis_compare}
	\vspace{-0.5em}
\end{figure*}

These supplementary materials provide additional analysis and visualizations that complement the main paper of \textbf{YOLOA: Real-Time Affordance Detection via LLM Adapter}. They are organized into five sections:
\begin{itemize}
	\setlength{\leftskip}{1em}
	\item Visualization of LLM Adapter effects.
	\item Evaluation of YOLOA with different LLMs.
	\item Hyperparameter study of $\alpha$, $\beta$, and $\gamma$.
	\item Analysis of top-$k$ prediction in affordance learning.
	\item Discussion of failure cases.
\end{itemize}

\section{Visualization of LLM Adapter Effects}

The visualization in Fig.~\ref{fig:vis_compare} highlights the crucial role of the LLM Adapter in improving object–affordance prediction. In the first row, the model without the LLM Adapter produces a ``display" activation that spills into the ``cup" region, resulting in an incorrectly located affordance mask relative to the ground truth. This mislocalized mask indicates that relying solely on visual cues is insufficient to distinguish semantically unrelated objects. With the LLM Adapter, the affordance mask is tightly confined to the ``tvm" regions, suggesting that the category- and location-aware guidance effectively suppresses spurious activations.

A similar issue arises in the second row: without the LLM Adapter, the model generates an affordance mask with inconsistent scale and even mixes affordance categories, undermining functional specificity. In contrast, the full configuration yields a well-scaled and clean ``contain" region that aligns precisely with the ``bowl", demonstrating that the LLM Adapter stabilizes affordance estimation by grounding it in object-level semantics.

In the third row, although the ground truth annotates two distinct ``pound" regions along the ``hammer" head, the model without the LLM Adapter produces weak and incomplete responses. After incorporating the LLM Adapter, both functional regions are properly recovered, exhibiting stronger activation intensity and improved discrimination. These observations collectively show that the LLM Adapter not only sharpens spatial localization but also enhances the semantic reliability of affordance prediction.

\begin{table}[t]
	\caption{\textbf{Evaluation of YOLOA with different LLMs on the ADG-Det dataset.} The LLM Adapter is tested with multiple LLM choices to assess its dependence on the underlying large language model. ($\uparrow$ / $\downarrow$  denote higher/lower is better.)}
	\label{tab:llm_backbone}
	\centering
	\resizebox{\columnwidth}{!}{\setlength{\heavyrulewidth}{1.5pt}
		\begin{tabular}{l@{\hspace{1pt}}cc@{\hspace{3pt}}c@{\hspace{3pt}}c@{\hspace{3pt}}c@{\hspace{3pt}}c}
			\toprule
			\multirow[c]{2}{*}{\vspace{-5pt}\textbf{LLM}} &
			\multirow[c]{2}{*}{\vspace{-5pt}\textbf{\#Params}} &
			\multicolumn{5}{c}{\textbf{Metrics}} \\
			\cmidrule(lr){3-7}
			& & \textbf{mAP}$\uparrow$ & \textbf{AR}$\uparrow$ & \textbf{KLD}$\downarrow$ & \textbf{SIM}$\uparrow$ & \textbf{NSS}$\uparrow$ \\
			\midrule
			w/o LLM &--& 49.6 & 43.7 & 1.812 & 0.293 & 1.176\\
			Phi-4-Mini & 3.8B & 51.4 & 47.9 & 1.727 & 0.320 & 1.235 \\
			LLaMA-3 & 8B & \textbf{52.8} & \textbf{49.7} & \textbf{1.528} & \textbf{0.341} & \textbf{1.346}\\
			\bottomrule
	\end{tabular}}
    \vspace{-1em}
\end{table}

\section{Evaluation of YOLOA with Different LLMs}

We further evaluate YOLOA by replacing the LLM employed in the LLM Adapter with different large language models to examine whether the effectiveness of the LLM Adapter depends on the model scale. As shown in Table~\ref{tab:llm_backbone}, the larger LLaMA-3 (8B)~\cite{touvron2023llama} achieves higher performance across all metrics. Notably, the smaller Phi-4-Mini (3.8B)~\cite{abdin2025phi4}, despite having less than half as many parameters as LLaMA-3, still yields substantial gains compared to the no-LLM baseline. This comparison reveals that even lightweight LLMs can provide coherent semantic cues for affordance reasoning, enabling the visual and linguistic features to be aligned within a shared latent space. Overall, we conclude that the effectiveness of the LLM Adapter is not heavily dependent on the model scale and can still be retained when using lightweight yet effective LLMs.

\section{\boldmath Hyperparameter Study on $\alpha, \beta, \gamma$}

\begin{table}[t]
	\centering
	\caption{\textbf{Hyperparameter study on $\alpha$, $\beta$, and $\gamma$ on the ADG-Det dataset.} Only one hyperparameter (\textbf{Hyper.}) is varied at a time, with the others fixed to the default value of $0.01$. }
	\label{tab:hyper_param}
	\resizebox{\columnwidth}{!}{
		\setlength{\heavyrulewidth}{1.5pt}
		\begin{tabular}{c@{\hspace{14pt}}cc@{\hspace{6pt}}c@{\hspace{6pt}}c@{\hspace{6pt}}c@{\hspace{6pt}}c}
			\toprule
			\multirow[c]{2}{*}{\vspace{-5pt}\textbf{Hyper.}} &
			\multirow[c]{2}{*}{\vspace{-5pt}\textbf{Value}} &
			\multicolumn{5}{c}{\textbf{Metrics}} \\
			\cmidrule(lr){3-7}
			& & \textbf{mAP}$\uparrow$ & \textbf{AR}$\uparrow$ & \textbf{KLD}$\downarrow$ & \textbf{SIM}$\uparrow$ & \textbf{NSS}$\uparrow$ \\
			\midrule
			
			\multirow{3}{*}{$\alpha$}
			& 0.001 & 51.7 & 47.3 & 1.620 & 0.329 & 1.288 \\
			& 0.01  & 52.6 & 49.0 & 1.612 & 0.330 & 1.294 \\
			& 0.1   & 52.3 & 48.5 & 1.625 & 0.335 & 1.286 \\
			\midrule
			
			\multirow{3}{*}{$\beta$}
			& 0.001 & 52.1 & 48.2 & 1.617 & 0.324 & 1.289 \\
			& 0.01  & 52.6 & 49.0 & 1.612 & 0.330 & 1.294 \\
			& 0.1   & 51.4 & 48.6 & 1.623 & 0.319 & 1.292 \\
			\midrule
			
			\multirow{3}{*}{$\gamma$}
			& 0.001 &\textbf{52.8} & \textbf{49.7} & \textbf{1.528} & \textbf{0.341} & \textbf{1.346} \\
			& 0.01  & 52.6 & 49.0 & 1.612 & 0.330 & 1.294 \\
			& 0.1   & 52.3 & 49.4 & 1.970 & 0.293 & 1.073 \\
			\bottomrule
	\end{tabular}}
    \vspace{-1em}
\end{table}
We further perform a hyperparameter study on the ADG-Det dataset to evaluate the impact of $\alpha$, $\beta$, and $\gamma$. We vary only one hyperparameter at a time, keeping the others fixed to the default value of $0.01$.:

\begin{itemize}
	\setlength{\leftskip}{1.5em}
	\item$\alpha$: class priors refinement weight 
	\item$\beta$: box offsets refinement weight  
	\item$\gamma$: affordance gates refinement weight
\end{itemize}

As presented in Table~\ref{tab:hyper_param}, both $\alpha$ and $\beta$ primarily affect object prediction (mAP, AR) by modulating the weights for category and location refinement. In contrast, $\gamma$ has a more significant impact on heatmap-based metrics, particularly KLD and SIM, as it controls the strength of affordance supervision, directly influencing the learning of object-affordance relationships. Notably, the best performance is achieved when $\alpha=\beta=0.01$, and $\gamma = 0.001$, consistent with the setup adopted in the main paper.

\section{Top-k prediction in Affordance Learning}

\begin{figure}[!t]
	\centering
	\includegraphics[width=\columnwidth]{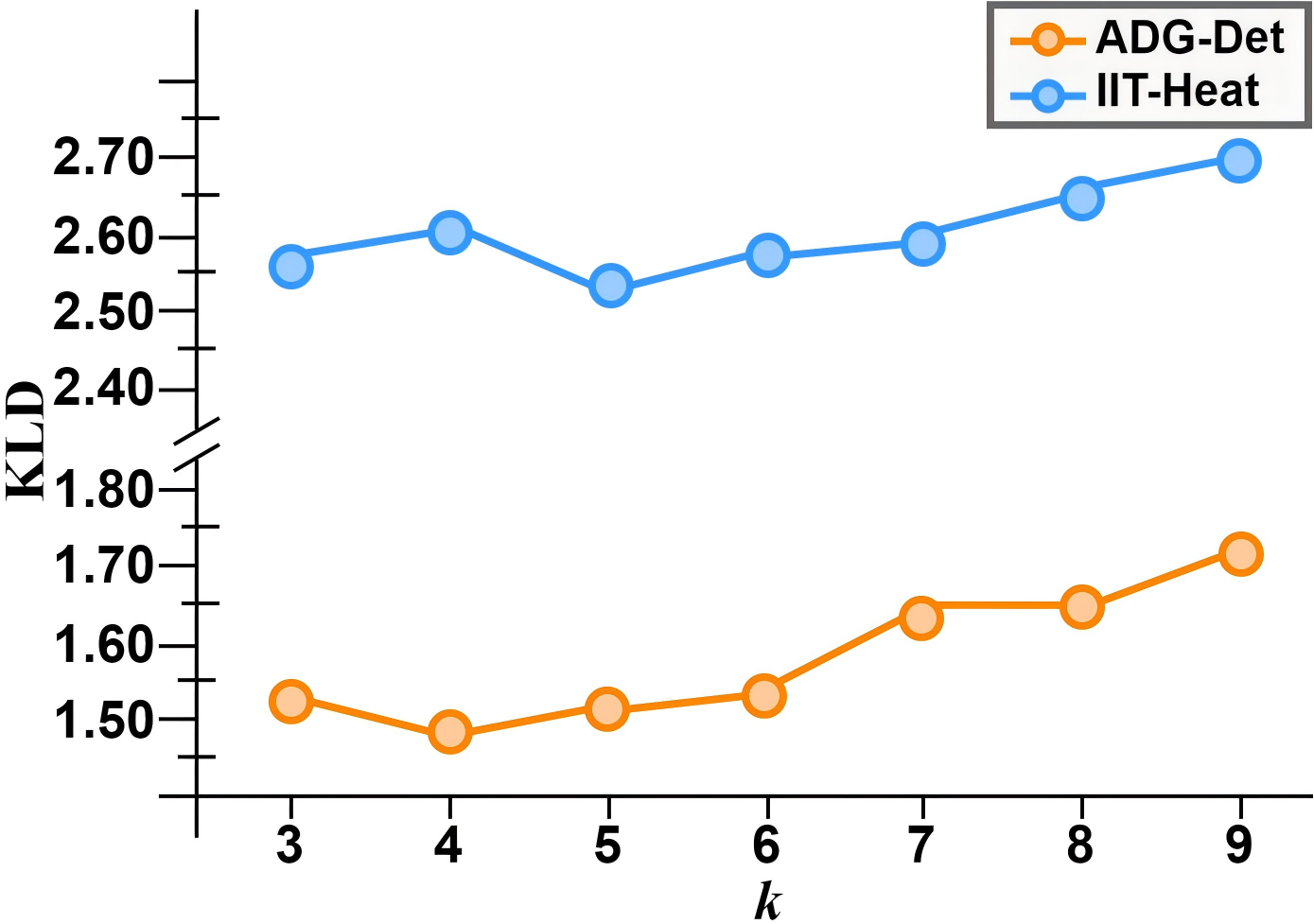}
	\caption{\textbf{Ablation study examining the effect of top-$k$ prediction on affordance learning.} Performance is measured using KLD on the ADG-Det and IIT-Heat datasets.}
	\label{fig:topk_KLD}
	\vspace{-1.5em}
\end{figure}

To analyze the effect of the top-$k$ prediction in affordance learning, we report the KLD results under different values of $k$. As shown in Fig.~\ref{fig:topk_KLD}, varying $k$ leads to minor variations in KLD, indicating that the affordance learning branch is relatively insensitive to the choice of $k$. This can be attributed to the fact that both datasets contain only a small number of objects per image; therefore, the top-$k$ predictions fed into the affordance learning branch provide similar category and location cues. Nevertheless, these cues remain essential for grounding affordance masks on the correct objects. The degradation observed when $k$ becomes excessively large suggests that overly large $k$ values introduce low-confidence object predictions, thereby harming the quality of affordance supervision. Based on these observations, we adopt the $k$ value (5 for ADG-Det, 6 for IIT-Heat) that yields the best object prediction performance in the main paper.

\section{Failure Case Discussion}

\begin{figure}[!t]
	\centering
	\includegraphics[width=\columnwidth]{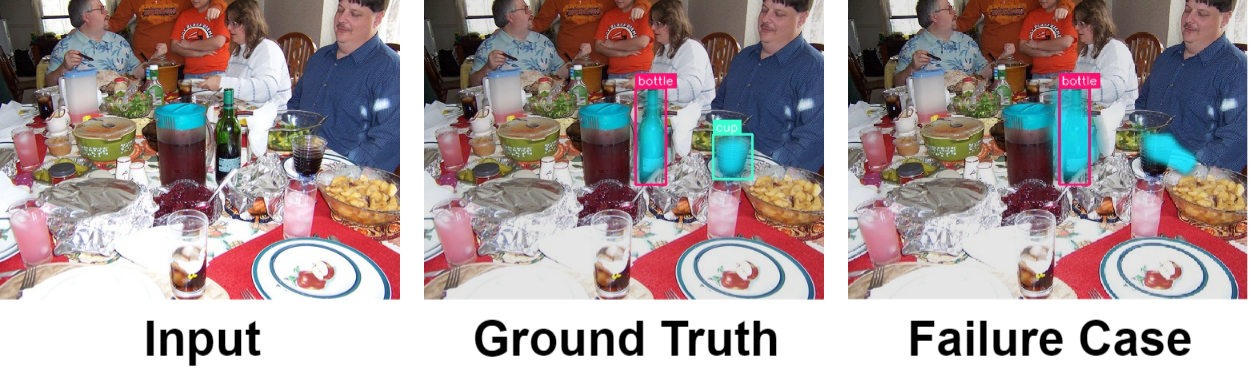}
	\caption{\textbf{Failure case on the IIT-Heat dataset.} Missing detections in real-world cluttered scenes lead to incomplete cues for the affordance learning branch.}
	\label{fig:failure}
	\vspace{-1em}
\end{figure}

In the IIT-Heat dataset, we note that YOLOA may fail in real-world cluttered scenes where object detection misses certain instances. As shown in Fig.~\ref{fig:failure}, when the object detection branch fails to detect an object, the affordance learning branch receives incomplete category and location cues, causing the predicted affordance mask to be incorrectly scaled and to spill beyond the corresponding functional region. This failure case illustrates the dependence of affordance prediction on accurate object detection and highlights the challenge of maintaining affordance quality under severe occlusion and visual clutter.

\section{Conclusion}

The supplementary experiments further validate the robustness and generality of YOLOA. The results demonstrate that the LLM Adapter consistently enhances object–affordance alignment, maintains stable performance across different LLM choices and hyperparameter settings, and remains reliable under various top-$k$ configurations. While failure cases highlight challenges in real-world cluttered scenes, the overall findings reinforce the effectiveness of the proposed design and its capacity to generalize across diverse affordance datasets.

\end{document}